\documentclass[conference]{IEEEtran}
\usepackage[namelimits]{amsmath} 
\usepackage{amssymb}             

\ifCLASSINFOpdf
\usepackage{graphicx}
 \usepackage{mathrsfs}
\else
\fi
\usepackage{array}

\usepackage{multirow}

\usepackage[justification=centering]{caption}
\usepackage{slashbox}
\usepackage[numbers,sort&compress]{natbib}
\usepackage[perpage,symbol]{footmisc}

\usepackage{caption}
\usepackage{balance}
\begin{document}
%
\title{Classification of Medical Images and Illustrations in the Biomedical Literature Using Synergic Deep Learning}

\author{\IEEEauthorblockN{${Jianpeng \ Zhang}^{1,2}$, ${Yong \ Xia}^{1,2*}$， ${Qi \ Wu}^{3}$, ${Yutong \ Xie}^{1,2}$}
\IEEEauthorblockA{1. Shaanxi Key Lab of Speech and Image Information Processing (SAIIP), \\School of Computer Science, Northwestern Polytechnical University, Xi'an 710072, China\\
2. Centre for Multidisciplinary Convergence Computing (CMCC), School of Computer Science, \\Northwestern Polytechnical University, Xi'an 710072, China\\
3. School of Computer Science, University of Adelaide, SA 5005, Australia\\
Email: yxia@nwpu.edu.cn}
}


%


\maketitle

\begin{abstract}
The Classification of medical images and illustrations in the literature aims to label a medical image according to the modality it was produced or label an illustration according to its production attributes. It is an essential and challenging research hotspot in the area of automated literature review, retrieval and mining. The significant intra-class variation and inter-class similarity caused by the diverse imaging modalities and various illustration types brings a great deal of difficulties to the problem. In this paper, we propose a synergic deep learning (SDL) model to address this issue. Specifically, a dual deep convolutional neural network with a synergic signal system is designed to mutually learn image representation. The synergic signal is used to verify whether the input image pair belongs to the same category and to give the corrective feedback if a synergic error exists. Our SDL model can be trained 'end to end'. In the test phase, the class label of an input can be predicted by averaging the likelihood probabilities obtained by two convolutional neural network components. Experimental results on the ImageCLEF2016 Subfigure Classification Challenge suggest that our proposed SDL model achieves the state-of-the-art performance in this medical image classification problem and its accuracy is higher than that of the first place solution on the Challenge leader board so far.
\end{abstract}

\textit{\textbf{Keywords---Medical image classification; Synergic deep learning model; Dual deep convolutional neural network}}

%
\IEEEpeerreviewmaketitle

\section{Introduction}
The indispensable role of digital medical imaging in the modern healthcare has led to the fast growth of digital images in all types of electronic biomedical publications. This fast growth poses great challenges for image retrieval, review and recruiting data for clinical care and research settings. Hence, there have been a large number of image classification researches aiming to improve the data mining ability in this area \cite{Ghosh:Review, Bruijne:Machine, Kalpathy-Cramer:Evaluating}. The ImageCLEF2016 Subfigure Classification Challenge \cite{Herrera:Overview}, recognizing the increasing complexity of images in biomedical literatures, contains figures with sub-figures that produced by multiple imaging modalities and illustrations drawn from analysis of medical data.

Image classification has been thoroughly studied during the past decades with a huge number of solutions being published in the literatures \cite{Li:A, Lazebnik:Beyond, Wang:Locality, Wang:Learning, Yang:A}. These solutions usually consist of the handcrafted feature extraction and classifier learning process. Despite of their success, it is difficult to design the handcrafted feature that is optimal for a specific classification task. Recently, with the deep learning methods being introduced, medical image analysis has experienced a rapid development. Especially, due to the fact that deep learning models can overcome the need for manual feature design and have superior classification capabilities, the medical image detection, classification \cite{Sirinukunwattana:Locality, Xie:Melanoma} and segmentation \cite{Chen:DCAN, Li:Deep} enjoyed a performance boosting. For example, Xu et al. \cite{Xu:Deep} adopted a deep convolutional neural network to minimize manual annotation and produce good feature representations for colon cancer classification using histopathology images. Shen et al. \cite{Shen:Multi} developed a multi-crop pooling strategy and applied it to a convolutional neural network to capture object salient information for lung nodule classification in CT images..

Although deep learning-based approaches outperform the state-of-the-art in a number of medical image analysis tasks in the field, substantial challenges still remain. For example, the issues with medical image datasets, including small datasets and anatomical variations, still restrict the effectiveness of classifying medical images and illustrations in the literature. The first issue usually relates to the work that required in acquiring the image data and then in image annotation \cite{Weese:Four}. Pre-trained deep convolutional neural network (DCNN) models have been used to address this issue, due to the strong transfer learning ability of the DCNN, i.e. learned from large-scale datasets like ImageNet, could be transferred to solving generic small-data visual recognition problems \cite{Oquab:Learning, Mettes:The}. Koitka et al. \cite{Koitka:Traditional} extracted the activation values from the last FC1000 layer in a pre-trained ResNet-152 model and adopted them to train a custom network layer using the pseudo inverse method \cite{Personnaz:Colloective}. Kumar et al. \cite{Kumar:An} proposed to integrate two different pre-trained CNN architectures and ensemble the results from multiple models into one high-quality classifier.

\begin{figure}[htb]
\centering
\includegraphics[height=1.5in]{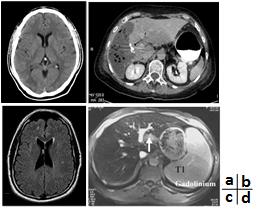}
\caption{An example shows the intra-class variation and inter-class similarity in modality-based medical image classification: (a) a brain CT image, (b) a pleural CT image, (c) a brain MR image, and (d) a pleural MR image.}
\end{figure}

\begin{figure*}[t]
\centering
\includegraphics[height=2.5in]{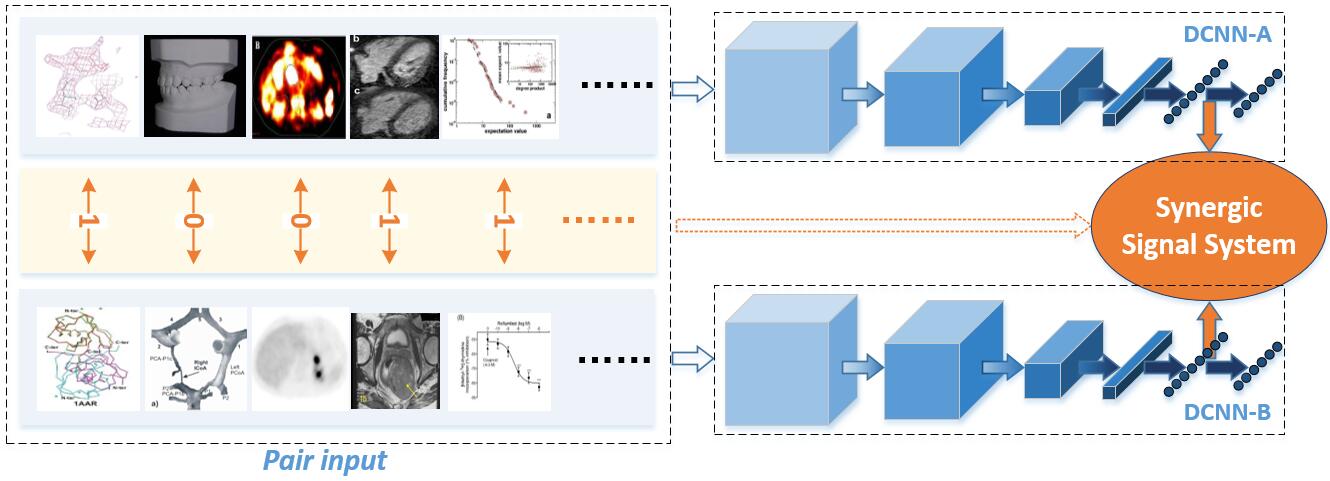}
\caption{Architecture of the SDL model, which consists of a dual deep convolutional neural network and an additional synergic signal system. The input is a pair of images which are sampled from the training set.}
\end{figure*}

The second issue is the intra-class variation and inter-class similarity \cite{Yang:Large}, which poses even greater challenges for classifying medical images according to the different modalities, by which they were produced. A typical example that highlights the difficulty is shown in Figure 1, in which the brain CT and pleural CT images in the top row looks dissimilar due to viewing different anatomical structures although they belong to the same category; whereas the brain CT and MRI CT images in the left column look very similar but belonging to different categories. Although deep neural networks have enough capacity to forcibly remember all training samples \cite{Zhang:Understanding}, the ambiguity produced by intra-class variation and inter-class similarity may tease a neural network and make it fall into confusion. The neural network makes right decision with the low confidence level and the results may even completely opposite if adding small fluctuations in the input.

In this paper, synergic deep learning (SDL) is presented to enhance the distinguishing ability of deep neural networks, especially for those confusing samples. The basic learning strategy of SDL is to use a synergic signal to bridge several neural networks so that they can guide and benefit from each other. We specifically design a SDL model that consists of a dual deep convolutional neural network (dual-DCNN) and a synergic signal system in this paper to solve the medical image classification problems. It's advisable to initialize our dual-DCNN with the pre-trained DCNN, whose parameters were derived from the ImageNet dataset \cite{Deng:ImageNet} and further fine tune it with our dataset. However, to break the independence between dual DCNNs, an additional synergic signal, serving as an information bridge between two DCNNs, is used to verify whether the input pair belongs to the same category or not. The wrong decision made by a DCNN will be highlighted in the form of synergic error with the help of correct decision made from another DCNN. In this case, the advantage depicted by one neural network is able to guide the learning of the weaker one, and both DCNNs have stronger representation ability to distinguish the confusing images with significant intra-class variance and inter-class similarity in the synergic learning mode. Moreover, our SDL model is easily trained under the classification and synergic supervisions in an 'end to end' fashion. In the test phase, the precision probabilities of each test sample given by dual neural networks are added together as an ensemble decision probability. We have evaluated it on the ImageCLEF2016 medical classification challenge dataset and the experimental results show that our proposed SDL model is the state-of-the-art on the medical classification problem.

\section{The Synergic Deep Learning Model}

Our proposed SDL model consists of three main components, i.e. a data pair input layer, a dual-DCNN and a synergic signal system, as shown in Figure 2. Different from the one by one input mode of conventional deep models, our SDL model accepts a pair of inputs that are randomly selected from the training set. A dual-DCNN, including DCNN-A and DCNN-B, is the main learning module with two input sequences. Both DCNN-A and DCNN-B are pre-trained residual neural networks and fine-tuned with the supervision of the true labels of inputting sequences. Besides, a synergic signal system is used to verify whether the input pair belongs to a same category or not, and gives the corrective feedback if a synergic error exists. For instance, in Figure 2, the first pair of images with high structure similarity actually belong to different classes. The second pair comes from the same class, but visually they are different. It's easy for a weak DCNN to make false decision under the contrast. The error generated from synergic signal system will further modify the dual-DCNN to have a stronger ability to distinguish these confusing samples.

We discuss the details of our proposed SDL model in the following sections.

\subsection{The Dual Deep Convolutional Neural Network}
The dual-DCNN is an important module in our SDL model that contains two complete learning units, namely DCNN-A and DCNN-B. In principle, any DCNN with arbitrary structure can be embedded in our SDL model. Here, due to the strong representation capability of the famous residual network \cite{He:Deep}, we employ a pre-trained residual neural network (ResNet-50, as shown in Figure 3) as the initialization of our DCNN-A and DCNN-B. It is composed of 50 learnable layers, and its parameters have been converged by training on the ImageNet dataset, for an image classification task. It's worth to note that two parameter sets of the DCNN-A and DCNN-B, denoted by $\theta^A$ and $\theta^B$, are not shared. To adapt the ResNet-50 model to our image dataset, we replace all fully connected ({\em FC}) layers with a {\em FC} layer of 1024 neurons ({\em FC1024-A/B}), a {\em FC} layer of {\em K} neurons ({\em K}-class classification) and a softmax layer, and then fine tune the parameters of ResNet-50 by using our own training data. The weights of new FC layers are initialized by uniform distribution {\em U(-0.05, 0.05)}. The cross-entropy loss function of each DCNN is defined as \begin{equation}
l(\theta )=\frac{1}{M}\sum_{i=1}^{M}[log(\sum_{j=1}^{K}e^{\theta _{j}^{T}x^{(i))}})-\theta _{y^{(i)}}^{T}x^{(i))}]
\end{equation} where {\em M} is the number of training data. The mini-batch stochastic gradient descent (mini-batch SGD) algorithm is used to optimize the $\theta$.

\begin{figure}[!htb]
\centering
\includegraphics[height=0.6in]{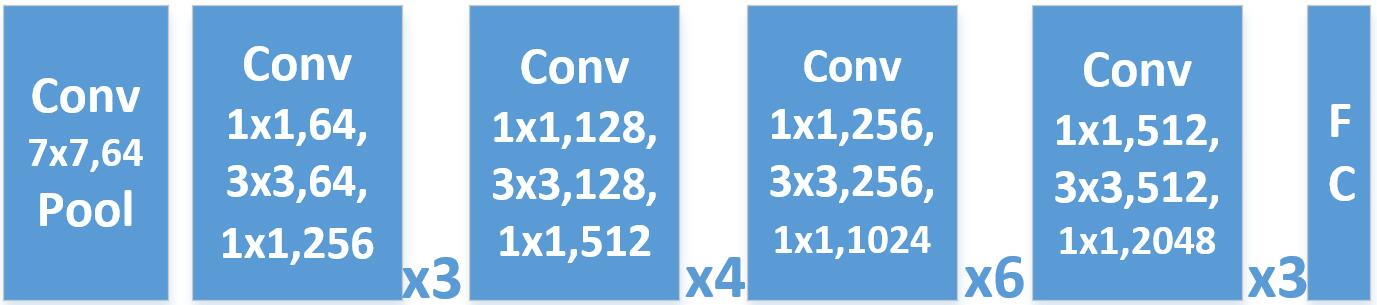}
\caption{Architecture of ResNet-50 model.}
\end{figure}

Both DCNN-A and DCNN-B accept input from a pair of images, aiming to supervise the training process in each learning unit with the true labels of corresponding input sequences. Although each DCNN has the ability to predict the label class of an input image, we creatively embed the activations from last two fully connected layers in both DCNNs into a synergic signal system to break the learning independence of the dual DCNNs.

\subsection{Synergic Signal System}
In our SDL model, a synergic signal system is designed to supervise the learning from the input pairs and bridge the gap between dual DCNNs. The architecture of this system is shown in Figure 4. Image representations need to be input into the synergic signal system in pairs. We randomly select image pairs from the training data and denote the property of a pair $(x_A, x_B)$ as
\begin{equation}S\left(x_{A},x_{B}\right) =\left\{
\begin{array}{l l}
1 & if\quad y_{A}=y_{B} \\
0 & if\quad y_{A}\neq y_{B}
\end{array}
\right..\end{equation} where $x_A$ and $x_B$ are outputs of {\em FC1024-A} and {\em FC1024-B}, $y_A$ and $y_B$ are true labels of $x_A$ and $x_B$, respectively. Here, `$S=1$' is a positive pair and `$S=0$' is a negative pair. Image pairs are selected from each min-batch. To avoid the unbalance data problem, the number of positive pairs in a batch is about $45\%-55\%$. $x_A$ and $x_B$ are concatenated together into an embedding layer followed by a {\em FC} layer with 2 neurons. It's convenient to monitor the synergic signal by adding another softmax layer and using the following cross-entropy loss
\begin{equation}
l^{S}(\theta ^{S})=\frac{1}{M}\sum_{i=1}^{M}{log(\sum_{j=1}^{K'}e^{{\theta ^{S}}_{j}^{T}(x_{A}^{T^{(i)}},x_{B}^{T^{(i)}})})-{\theta_{S}}_{y_{S}^{(i)}}^{T}(x_{A}^{T^{(i)}},x_{B}^{T^{(i)}})}
\end{equation} where $\theta^S$ is the ensemble of parameters of the synergic signal system. The detailed learning process of our proposed model is summarized in Table 1.

\begin{figure}[!htb]
\centering
\includegraphics[height=3in]{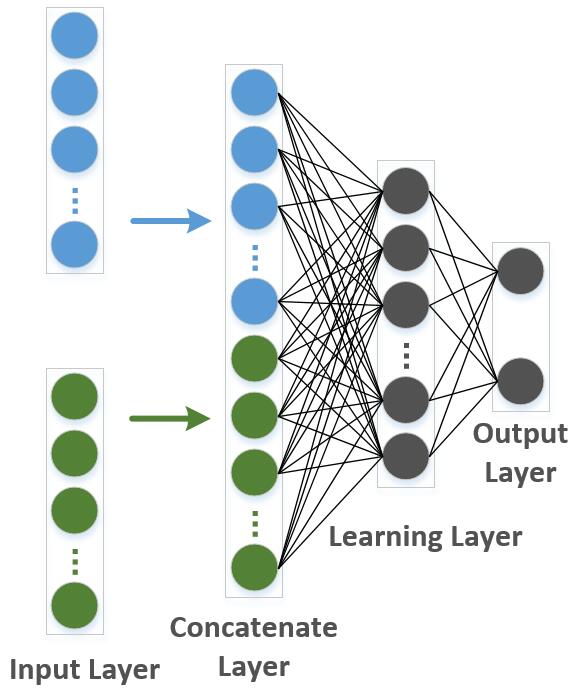}
\caption{Diagram of the synergic signal system.}
\end{figure}

\begin{table*}[htbp]
\centering
\caption{Learning process of the synergic deep learning model.}
\begin{tabular}{p{1cm} p{15cm}}
\hline
\textbf{Input}: & $x_A$ and $x_B$ (outputs of {\em FC1024-A} and {\em FC1024-B}), initialized parameters of DCNN-A, DCNN-B and the synergic signal system, $\theta^A$, $\theta^B$ and $\theta^S$, learning rate $\eta(t)$ and the hyper parameter $\lambda$. \\
\hline
Step1: & Concatenate two input features {\em FC1024-A} and {\em FC1024-B} into a combined one, which is denoted as $(x_A^T, x_B^T)$. The labels of these three supervisions are $y_A$, $y_B$, $y_S$. \\
Step2: & Update parameters $\theta^A$, $\theta^B$ and $\theta^S$ by using back-propagation algorithm. \\
 & Compute loss:
                $l^A(\theta^A)$, $l^B(\theta^B)$ and $l^S(\theta^S)$. \\
 & Compute gradient: \\
 & \setlength{\parindent}{10em} $\frac{\partial l^{A}(\theta^{A})}{\partial {\theta^{A}}_k}=\frac{1}{m}\sum_{i=1}^{m}{{x_A}^{(i)}[\frac{e^{{\theta^A}_{k}^{T}x_{A}^{(i)}}}{\sum_{j=1}^{K}e^{{\theta^A}_{j}^{T}x_{A}^{(i)}}}-\delta _{ky_{A}^{(i)}}]}, $ \\
 & \setlength{\parindent}{10em} $\frac{\partial l^{B}(\theta^{B})}{\partial {\theta^{B}}_k}=\frac{1}{m}\sum_{i=1}^{m}{{x_B}^{(i)}[\frac{e^{{\theta^B}_{k}^{T}x_{B}^{(i)}}}{\sum_{j=1}^{K}e^{{\theta^B}_{j}^{T}x_{B}^{(i)}}}-\delta _{ky_{B}^{(i)}}]}, $ \\
 & \setlength{\parindent}{10em} $\frac{\partial l^{S}(\theta^{S})}{\partial {\theta^{S}}_k'}=\frac{1}{m}\sum_{i=1}^{m}{{({x_A}^{T},{x_B}^{T})}^{(i)}[\frac{e^{{\theta^S}_{k'}^{T}({x_A}^{T},{x_B}^{T})^{(i)}}}{\sum_{j=1}^{K'}e^{{\theta^S}_{j}^{T}({x_A}^{T},{x_B}^{T})^{(i)}}}-\delta _{ky_{S}^{(i)}}]}, $ \\
 & \setlength{\parindent}{10em} $\Delta ^{A}=\frac{\partial l^{A}(\theta^{A})}{\partial {\theta^{A}}_k}+\lambda \frac{\partial l^{S}(\theta^{S})}{\partial {\theta^{S}}_{k'}}, \Delta ^{B}=\frac{\partial l^{B}(\theta^{B})}{\partial {\theta^{B}}_k}+\lambda \frac{\partial l^{S}(\theta^{S})}{\partial {\theta^{S}}_{k'}}, $ \\
 & \setlength{\parindent}{10em} where $\delta _{ky^(i)}=\left\{\begin{array}{l l}
                      1 & k=y^{(i)}\\
                      0 & k\neq y^{(i)}
                      \end{array}\right.$ and $\lambda$ is a weighting factor of synergic signal. \\
 & Update parameters:
                $\theta^A=\theta^A-\eta(t)\cdot \Delta^A$ and $\theta^B=\theta^B-\eta(t)\cdot \Delta^B$\\

\hline
\end{tabular}
\end{table*}

\subsection{Test Phase}
In the test phase, for a test image {\em x}, the DCNN-A and DCNN-B give predictions $P^A=(p_1^A,p_2^A,...,p_K^A)$ and $P^B=(p_1^B,p_2^B,...,p_K^B)$, which are activations in last {\em FC} layer. The additional synergic signal is abandoned for final classification results in the test phase. The predicted label of the input {\em x} is denoted as 
\begin{equation}
\underset{i}{argmax}\{(p_1^A+p_1^B),...,(p_i^A+p_i^B),...,(p_K^A+p_K^B)\}
\end{equation}

\section{Experiments}
\subsection{A toy example}
In this section, a toy example on the MNIST dataset is presented. Based on the classical LeNet-5 model, we designed a simple convolutional neural network called LeCNN, shown in Figure 5, as the architecture of DCNN-A and DCNN-B modules. Thus, the SDL model is composed of two LeCNNs and a synergic signal system. To simply evaluate the effectiveness of our SDL model, we set the $\lambda$=1. Figure 6 shows the loss and accuracy curves obtained on the training and testing sets during the training progress. It suggests that using the synergic signal system does lead to performance improvement, as our SDL model has lower loss and higher accuracy than the LeCNN model.

\begin{figure}
\centering
\includegraphics[height=0.55in]{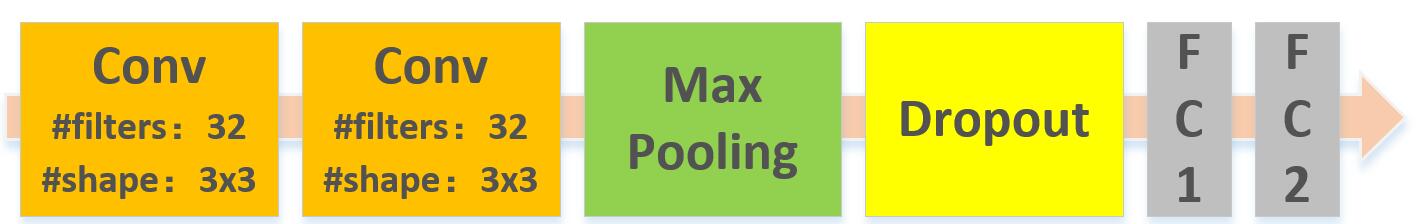}
\caption{A simple convolutional neural network (LeCNN) architecture used in MNIST experiment.}
\end{figure}

\begin{figure}
\centering
\includegraphics[height=2.5in]{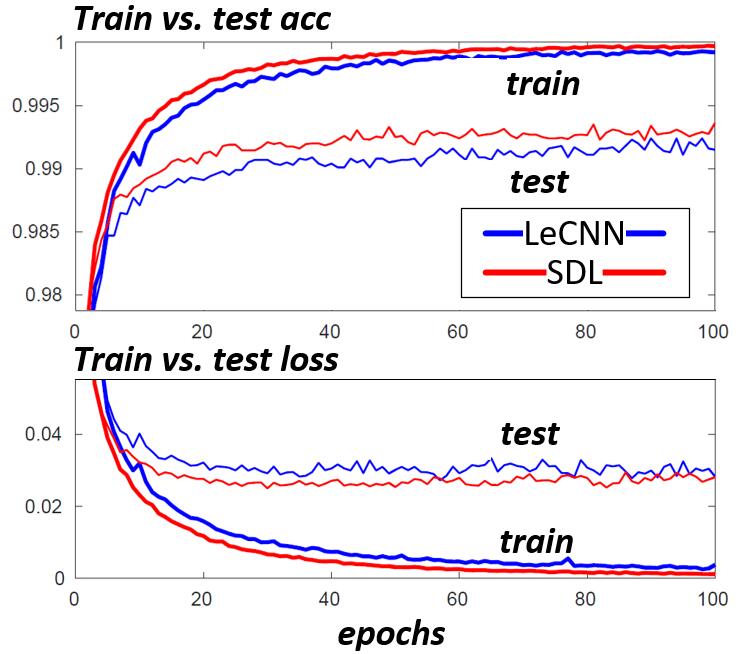}
\caption{Comparison of LeCNN and proposed SDL on the MNIST dataset.}
\end{figure}

\subsection{Dataset}
We evaluated our SDL model on the ImageCLEF2016 Subfigure Classification Challenge dataset \cite{Herrera:Overview}, which consists of 6776 training images and 4166 testing images collected from the PubMed Central (PMC\footnote{http://www.ncbi.nlm.nih.gov/pmc/})\cite{Mueller:Creating}. These images are divided into 30 categories, including 12 categories of medical diagnostic images, such as CT images, MRI images and PET images, and 18 categories of illustrations, such as figures, tables and flow charts. The abbreviations and details of each image category were listed in Table 2. The aim of our experiment is to classify medical diagnostic images according to the modality they were produced and classify illustrations according to their production attributes.

\begin{table}[t]
\centering
\caption{Category abbreviations and details in the ImageCLEF2016 classification hierarchy.}
\begin{tabular}{cccc}
\hline
No. & Abb. & Det. & Training \\
\hline
1&D3DR&3D reconstructions&201 \\
2&DMEL&Electron microscopy&208 \\
3&DMFL&Fluorescence microscopy&906 \\
4&DMLI&Light microscopy&696 \\
5&DMTR&	Transmission microscopy&300 \\
6&DRAN&	Angiography&17 \\
7&DRCO&	Combined modalities in one image&33 \\
8&DRCT&	Computerized Tomography&61 \\
9&DRMR&	Magnetic Resonance&139 \\
10&DRPE&PET&14 \\
11&DRUS&Ultrasound&26 \\
12&DRXR&X-ray, 2D Radiography&51 \\
13&DSEC&Electrocardiography&10 \\
14&DSEE&Electroencephalography&8 \\
15&DSEM&Electromyography&5 \\
16&DVDM&Dermatology, skin&29 \\
17&DVEN&Endoscopy&16 \\
18&DVOR&Other organs&55 \\
19&GCHE&Chemical structure&61 \\
20&GFIG&Statistical figures, graphs, charts&2954 \\
21&GFLO&Flowcharts&20 \\
22&GGEL&Chromatography, Gel&344 \\
23&GGEN&Gene sequence&179 \\
24&GHDR&Hand-drawn sketches&136 \\
25&GMAT&Mathematics, formulate&15 \\
26&GNCP&Non-clinical photos&88 \\
27&GPLI&Program listing&1 \\
28&GSCR&Screenshots&33 \\
29&GSYS&System overviews&91 \\
30&GTAB&Tables and forms&79 \\
\hline\end{tabular}
\end{table}

\subsection{Parameter Settings}
To leverage the overfitting issue in deep learning, we utilized several data argumentation strategies, including rotation, translation and random scaling, to enlarge our dataset 10 times. We designed the following variable learning rate
\begin{equation}\eta (t)=\frac{\eta(0)}{1+10^{-4}\times t}\end{equation} where {\em t} is the index of iteration and $\eta(0)=0.00005$. We set the maximum epoch number to 80 and adopted the mini-batch stochastic gradient decent with a batch size 64 as the optimizer. To stop the training process when the model falls into overfitting, {\em 20\%} of training data were randomly selected to form a validation set, which was used to monitor the performance of our model. We evaluated our proposed model with different values of the hyper parameter $\lambda$. The performance shown in Figure 7 reveals that our model achieves the lowest loss and highest accuracy when $\lambda=40$. Hence, we empirically set the value of $\lambda$ to 40 in our experiments.

\begin{figure}[t]
\centering
\includegraphics[height=2.9in]{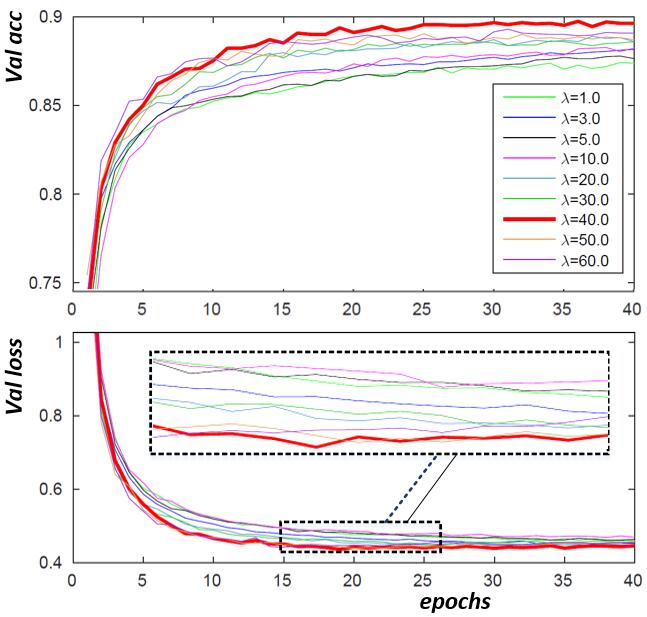}
\caption{Comparison of validation loss and accuracy under different $\lambda$ values.}
\end{figure}

\subsection{Results and Analysis}
We evaluated our proposed SDL model against the standard ResNet-50 model with the same experiment settings, including the same training set, validation set, initial parameters and learning rate scheme. Figure 8 shows the loss and accuracy curves of both models on the validation dataset. The smaller loss and higher accuracy achieved by the SDL model indicate that our model outperforms ResNet-50 on the validation set.

\begin{figure}[!htb]
\centering
\includegraphics[height=2.6in]{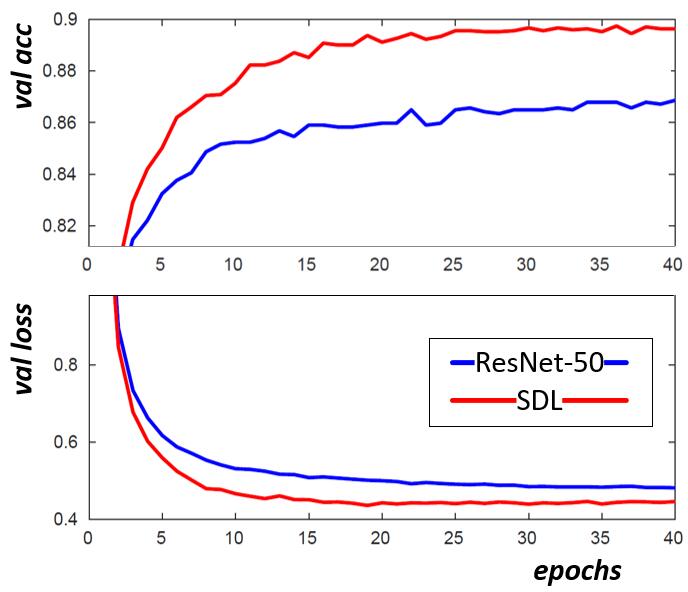}
\caption{Validation loss and accuracy curves by using pre-trained ResNet-50 and our proposed SDL model.}
\end{figure}

The classification accuracies of the ResNet-50 model, each component of our model i.e. DCNN-A or DCNN-B, and our SDL model on the validation set were displayed in Figure 9. It shows that, after incorporating the synergic signal system into the dual-DCNN architecture, each component of our model, which is also ResNet-50, achieves more than {\em 2\%} accuracy improvement, as compared to the standard ResNet-50. Moreover, jointly using those two component in an ensemble learning manner can further improve the classification accuracy.

\begin{figure}[!htb]
\centering
\includegraphics[height=1.55in]{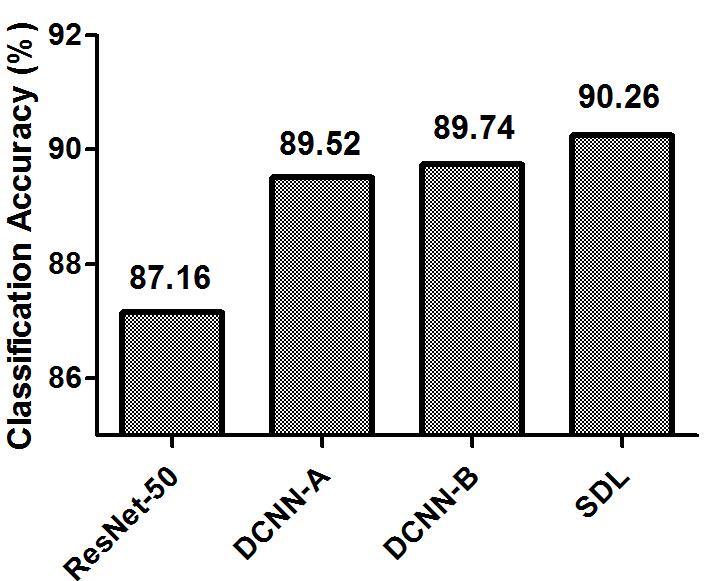}
\caption{Classification accuracy of ResNet-50, each DCNN component of our model and our SDL model on the validation set.}
\end{figure}

The F-scores \cite{Powers:Evaluation} of our SDL model and ResNet-50 model were calculated on each category of the test dataset and were depicted in Figure 10, in which a red arrow indicates an increased accuracy when applying our model to that category of test data, whereas a blue arrow suggests a decreased accuracy. It shows that our model achieves higher classification accuracy than ResNet-50 on most categories.

\begin{figure}[!htb]
\centering
\includegraphics[height=4in]{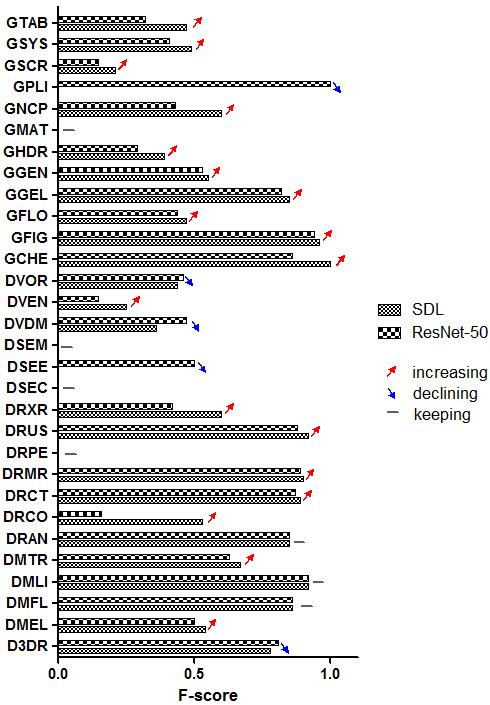}
\caption{Comparison of F-scores for each test class using ResNet-50 and our SDL model.}
\end{figure}

Next, we evaluated our SDL model on the ImageCLEF2016 test set. Figure 11 gives the confusion matrix of the classification results obtained by using our SDL model. The x-axis is the predicted label, and y-axis is the true label. A higher intensity value represents higher classification accuracy. Since this is a highly imbalanced classification problem, we let the size of each rectangle in this figure be proportional to the number of training images in each class. The confusion matrix shows that our SDL model achieves relatively accurate classification on every major category and most minor categories.

\begin{figure}[h]
\centering
\includegraphics[height=2.5in]{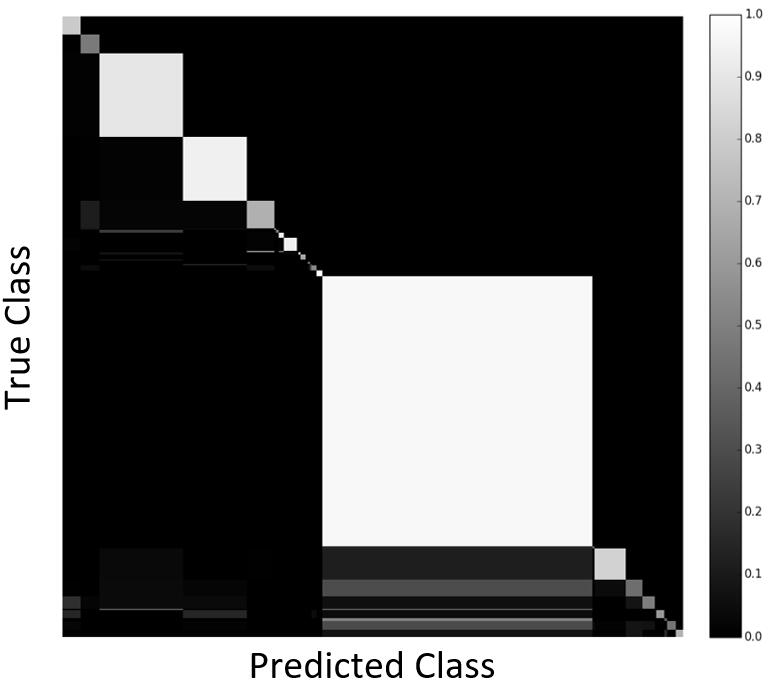}
\caption{Confusion matrix of the classification result obtained by applying our SDL model to the ImageCLEF2016 test dataset.}
\end{figure}

Table 3 gives the classification accuracy of our proposed SDL model, ResNet-50 model, Kumar's ensemble method and the accuracy of six best-performed solutions listed in the ImageCLEF2016 Subfigure Classification Challenge leader board. Obviously, handcraft feature engineering underperforms deep learning-based methods, and deeper network, such as ResNet-152, performs better than ResNet-50. However, our SDL model achieves the so far best classification accuracy in this challenge. Please note that our SDL model is capable of adapting any DCNN structures, which means it can benefit from much deeper models, such as ResNet-152.

\begin{table}[t]
\centering
\caption{Classification accuracy of our SDL model and eight best-performed solutions on the ImageCLEF2016 test dataset.}
\begin{tabular}{p{6cm} p{2cm}}
\hline
Method&Classification Accuracy(\%)\\
\hline
\textbf{SDL model} & \textbf{86.58} \\
Koitka et.al \cite{Koitka:Traditional} (ResNet-152) & 85.38 \\
ResNet-50 & 84.54 \\
Koitka et.al \cite{Koitka:Traditional} (11 handcrafted features) & 84.46 \\
Valavanis et.al \cite{Valavanis:IPL} (BoW model) & 84.01 \\
Kumar et.al\cite{Kumar:An}(Ensemble of multi-DCNNs) & 82.48 \\
Kumar et.al \cite{Kumar:Subfigure} (A pre-trained DCNN) & 77.55 \\
Li et.al \cite{Li:UDEL} & 72.46 \\
Semedo et.al \cite{Semedo:NovaSearch} & 65.31 \\
\hline\end{tabular}
\end{table}

\section{Conclusions}
In this paper, we propose a synergic deep learning (SDL) model that contains a dual collaborative deep convolutional neural network and an additional synergic signal system to classify medical images and illustrations in the biomedical literature. To strength the collaborative learning of a dual nets, the synergic signal is used to verify whether an input pair belongs to a same category or not. It promotes that our SDL model has stronger representation ability to distinguish those easily confused inter-class samples and obvious diversity of intra-class samples. Experimental results on the ImageCLEF2016 Subfigure Classification Challenge dataset show that our proposed SDL model achieves the state-of-the-art performance in this medical image classification problem, with an accuracy higher than the first place on the Challenge leader board at the time of submission.

\section{Acknowledgements}
We appreciate the efforts devoted by the organizers of the ImageCLEF2016 Subfigure Classification Challenge to collect and share the data for comparing algorithms of classifying medical images and illustrations in the biomedical literature.




\bibliographystyle{abbrv}
\bibliography{sigproc}

\begin{thebibliography}{10}

\bibitem{Chen:DCAN}
H.~Chen, X.~Qi, L.~Yu, Q.~Dou, J.~Qin, and P.-A. Heng.
\newblock Dcan: Deep contour-aware networks for object instance segmentation
  from histology images.
\newblock {\em Medical Image Analysis}, 36:135--146, Feb 2017.

\bibitem{Bruijne:Machine}
M.~de~Bruijne.
\newblock Machine learning approaches in medical image analysis: From detection
  to diagnosis.
\newblock {\em Medical Image Analysis}, 33:94--97, Oct 2016.

\bibitem{Herrera:Overview}
A.~G.~S. de~Herrera, R.~Schaer, S.~Bromuri, and H.~Mueller.
\newblock Overview of the imageclef 2016 medical task.
\newblock In {\em CLEF2016 Working Notes. CEUR Workshop Proceedings}, Sep 2016.

\bibitem{Deng:ImageNet}
J.~Deng, W.~Dong, R.~Socher, L.-J. Li, K.~Li, and F.-F. Li.
\newblock Imagenet: A large-scale hierarchical image database.
\newblock In {\em IEEE Conference on Computer Vision and Pattern Recognition},
  pages 248--255, Jun 2009.

\bibitem{Ghosh:Review}
P.~Ghosh, S.~Antani, L.~R. Long, and G.~R. Thoma.
\newblock Review of medical image retrieval systems and future directions.
\newblock In {\em 24th IEEE International Symposium on Computer-Based Medical
  Systems}, Jun 2011.

\bibitem{He:Deep}
K.~He, X.~Zhang, S.~Ren, and J.~Sun.
\newblock Deep residual learning for image recognition.
\newblock In {\em IEEE Conference on Computer Vision and Pattern Recognition},
  pages 770--778, June 2016.

\bibitem{Kalpathy-Cramer:Evaluating}
J.~Kalpathy-Cramer, A.~G.~S. de~Herrera, D.~Demner-Fushman, S.~Antani,
  S.~Bedrick, and H.~Mueller.
\newblock Evaluating performance of biomedical image retrieval systems an
  overview of the medical image retrieval task at imageclef 2004-2013.
\newblock {\em Computerized Medical Imaging And Graphics}, 39:55--61, Jan 2015.

\bibitem{Koitka:Traditional}
S.~Koitka and C.~M. Friedrich.
\newblock Traditional feature engineering and deep learning approaches at
  medical classification task of imageclef 2016 fhdo biomedical computer
  science group (bcsg).
\newblock In {\em CLEF2016 Working Notes. CEUR Workshop Proceedings}, Seq 2016.

\bibitem{Kumar:An}
A.~Kumar, J.~Kim, D.~Lyndon, M.~Fulham, and D.~Feng.
\newblock An ensemble of fine-tuned convolutional neural networks for medical
  image classification.
\newblock {\em IEEE Journal of Biomedical and Health Informatics},
  21(1):31--40, Jan 2017.

\bibitem{Kumar:Subfigure}
A.~Kumar, D.~Lyndon, J.~Kim, and D.~Feng.
\newblock Subfigure and multi-label classification using a fine-tuned
  convolutional neural network.
\newblock In {\em CLEF2016 Working Notes. CEUR Workshop Proceedings}, Sep 2016.

\bibitem{Lazebnik:Beyond}
S.~Lazebnik, C.~Schmid, and J.~Ponce.
\newblock Beyond bags of features: Spatial pyramid matching for recognizing
  natural scene categories.
\newblock In {\em IEEE Conference on Computer Vision and Pattern Recognition},
  pages 2169--2178, Jun 2006.

\bibitem{Li:A}
F.-F. Li and P.~Perona.
\newblock A bayesian hierarchical model for learning natural scene categories.
\newblock In {\em IEEE Conference on Computer Vision and Pattern Recognition},
  pages 524--531, Jun 2005.

\bibitem{Li:UDEL}
P.~Li, S.~Sorensen, A.~Kolagunda, X.~Jiang, X.~Wang, C.~Kambhamettu, and
  H.~Shatkay.
\newblock Udel cis at imageclef medical task 2016.
\newblock In {\em CLEF2016 Working Notes. CEUR Workshop Proceedings}, Sep 2016.

\bibitem{Li:Deep}
R.~Li, T.~Zeng, H.~Peng, and S.~Ji.
\newblock Deep learning segmentation of optical microscopy images improves 3d
  neuron reconstruction.
\newblock {\em IEEE Transactions on Medical Imaging}, PP, Mar 2017.

\bibitem{Mettes:The}
P.~Mettes, D.~C. Koelma, and C.~G.~M. Snoek.
\newblock The imagenet shuffle: Reorganized pre-training for video event
  detection.
\newblock In {\em ACM International Conference on Multimedia Retrieval}, pages
  175--182, Jun 2016.

\bibitem{Mueller:Creating}
H.~Mueller, J.~Kalpathy-Cramer, D.~Demner-Fushman, and S.~Antani.
\newblock Creating a classification of image types in the medical literature
  for visual categorization.
\newblock In {\em Conference on Medical Imaging - Advanced PACS-Based Imaging
  Informatics and Therapeutic Applications}, Feb 2012.

\bibitem{Oquab:Learning}
M.~Oquab, L.~Bottou, I.~Laptev, and J.~Sivic.
\newblock Learning and transferring mid-level image representations using
  convolutional neural networks.
\newblock In {\em IEEE Conference on Computer Vision and Pattern Recognition},
  pages 1717--1724, Jun 2014.

\bibitem{Personnaz:Colloective}
L.~Personnaz, I.~Guyon, and G.~Dreyfus.
\newblock Collective computational properties of neural networks: New learning
  mechanisms.
\newblock {\em Physical Review A}, 34:4217--4228, Nov 1986.

\bibitem{Powers:Evaluation}
D.~M.~W. Powers.
\newblock Evaluation: From precision, recall and f-factor to roc, informedness,
  markedness and correlation.
\newblock {\em International Journal of Machine Learning Technology},
  2(1):37--63, Dec 2011.

\bibitem{Semedo:NovaSearch}
D.~Semedo and J.~Magalhaes.
\newblock Novasearch at imageclefmed2016 subfigure classification task.
\newblock In {\em CLEF2016 Working Notes. CEUR Workshop Proceedings}, Sep 2016.

\bibitem{Shen:Multi}
W.~Shen, M.~Zhou, F.~Yang, D.~Yu, D.~Dong, C.~Yang, Y.~Zang, and J.~Tian.
\newblock Multi-crop convolutional neural networks for lung nodule malignancy
  suspiciousness classification.
\newblock {\em Pattern Recognition}, 61:663--673, Jan 2017.

\bibitem{Sirinukunwattana:Locality}
K.~Sirinukunwattana and S.~Raza.
\newblock Locality sensitive deep learning for detection and classification of
  nuclei in routine colon cancer histology images.
\newblock {\em IEEE Transactions on Medical Imaging}, 35(5):1196--1206, May
  2016.

\bibitem{Valavanis:IPL}
L.~Valavanis, S.~Stathopoulos, and T.~Kalamboukis.
\newblock Ipl at clef2016 medical task.
\newblock In {\em CLEF2016 Working Notes. CEUR Workshop Proceedings}, Sep 2016.

\bibitem{Wang:Locality}
J.~Wang, J.~Yang, K.~Yu, F.~Lv, T.~Huang, and Y.~Gong.
\newblock Locality-constrained linear coding for image classification.
\newblock In {\em IEEE Conference on Computer Vision and Pattern Recognition},
  pages 3360--3367, Jun 2010.

\bibitem{Wang:Learning}
Z.~Wang, Y.~Hu, and L.-T. Chia.
\newblock Learning image-to-class distance matric for image classification.
\newblock {\em ACM Tansactions on Intelligent Systems and Technology}, 4(2),
  Mar 2013.

\bibitem{Weese:Four}
J.~Weese and C.~Lorenz.
\newblock Four challenges in medical image analysis from an industrial
  perspective.
\newblock {\em Medical Image Analysis}, 33:44--49, Oct 2016.

\bibitem{Xie:Melanoma}
F.~Xie, H.~Fan, and Y.~Li.
\newblock Melanoma classification on dermoscopy images using a neural network
  ensemble mode.
\newblock {\em IEEE Transactions on Medical Imaging}, 36(3):849--858, Mar 2017.

\bibitem{Xu:Deep}
Y.~Xu, T.~Mo, Q.~Feng, P.~Zhong, M.~Lai, and E.~I.-C. Chang.
\newblock Deep learning of feature representation with multiple instance
  learning for medical image analysis.
\newblock In {\em IEEE International Conference on Acoustics, Speech and Signal
  Processing}, pages 1626--1630, Jul 2014.

\bibitem{Yang:Large}
S.~Yang, W.~Cai, H.~Huang, Y.~Zhou, D.~D. Feng, Y.~Wang, M.~J. Fulham, and
  M.~Chen.
\newblock Large margin local estimate with applications to medical image
  classification.
\newblock {\em IEEE Transactions on Medical Imaging}, 34(6):1362--1377, Jun
  2015.

\bibitem{Yang:A}
Y.~Yang, L.~Yang, G.~Wu, and S.~Li.
\newblock A bag-of-objects retrieval model for web image search.
\newblock In {\em ACM International Conference on Multimedia}, pages 49--58,
  Oct 2012.

\bibitem{Zhang:Understanding}
C.~Zhang, S.~Bengio, M.~Hardt, and O.~Vinyals.
\newblock Understanding deep learning requires rethinking generalization.
\newblock In {\em International Conference on Learning Representations}, Apr
  2017.

\end{thebibliography}

\end{document}